\documentclass[letterpaper, 10 pt, conference]{ieeeconf}  

\IEEEoverridecommandlockouts                              

\usepackage{graphics} 
\usepackage{epsfig} 
\usepackage{times} 
\usepackage{amsmath} 
\usepackage{amssymb}  
\usepackage{bm}
\usepackage[linesnumbered,ruled,noend]{algorithm2e} 
\usepackage[noend]{algpseudocode} 
\usepackage{caption}
\usepackage{subcaption}

\usepackage[backend=bibtex,
            hyperref=true,
            url=false,
            isbn=false,
            doi=false,
            backref=false,
            style=ieee,
            natbib=true,
            mincitenames=1,
            maxcitenames=1,
            citestyle=numeric-comp,
            sorting=nyt,
            block=none]{biblatex}

\newcommand{\newsec}[1]{\vspace{2mm} \noindent \textbf{#1.} }

\usepackage{xspace}
\newcommand{\alg}{MoP-RL\xspace}
\newcommand{\alglong}{Model-based Preference-based Reinforcement Learning\xspace}

\usepackage{hyperref}
\usepackage{color}

\title{\LARGE \bf
Efficient Preference-Based Reinforcement Learning \\
Using Learned Dynamics Models
}

\author{Yi Liu$^{1}$, Gaurav Datta$^{1}$, Ellen Novoseller$^{2}$, Daniel S. Brown$^{3}$
\thanks{$^{1}$UC Berkeley,
  $^{2}$Army Research Lab, $^{3}$University of Utah
    }%
}

\begin{document}

\maketitle
\thispagestyle{empty}
\pagestyle{empty}

\begin{abstract}

Preference-based reinforcement learning (PbRL) can enable robots to learn to perform tasks based on an individual's preferences without requiring a hand-crafted reward function. However, existing approaches either assume access to a high-fidelity simulator or analytic model or take a model-free approach that requires extensive, possibly unsafe online environment interactions. In this paper, we study the benefits and challenges of using a \textit{learned dynamics model} when performing PbRL. In particular, we provide evidence that a learned dynamics model offers the following benefits when performing PbRL: (1) preference elicitation and policy optimization require significantly fewer environment interactions than model-free PbRL, (2) diverse preference queries can be synthesized safely and efficiently as a byproduct of standard model-based RL, and (3) reward pre-training based on suboptimal demonstrations can be performed without any environmental interaction. Our paper provides empirical evidence that learned dynamics models enable robots to learn customized policies based on user preferences in ways that are safer and more sample efficient than prior preference learning approaches. Supplementary materials and code are available at \url{https://sites.google.com/berkeley.edu/mop-rl}.

\end{abstract}

\section{INTRODUCTION}

Developing assistive and collaborative robots that adapt to a variety of end-users requires customizing robot behaviors without manually tuning cost functions. One common approach is for robots to learn from human demonstrations~\cite{argall2009survey,hussein2017imitation,ravichandar2020recent}. However, providing demonstrations is not always possible and when demonstrations are suboptimal existing methods often overfit to the suboptimalities~\cite{chuck2017statistical,mandlekar2022matters,gopalannegative,ghosal2023effect}. 

Recent years have seen progress in preference-based reinforcement learning (PbRL)---an approach that, rather than relying on demonstrations, queries the user for pairwise preferences over trajectories~\cite{wirth2017survey}. However, existing PbRL approaches either use model-free PbRL and assume access to a perfect simulator~\cite{christiano2017deep,ibarz2018reward,lee2021pebble} or or access to an analytic dynamics model~\cite{sadigh2017active}. In many problems, however, a good dynamics model is not available and must be inferred from data, as done in model-based RL~\cite{chua2018deep}. Prior work on PbRL has not addressed the challenges and potential benefits of using a learned dynamics model to perform PbRL. Due to the importance of model-based RL in robotics domains, we propose to study Model-based Preference-based RL (MoP-RL). To our knowledge, this work is the first to study the benefits of learning a dynamics model in preference learning for realistic domains such as robotics. Central to our paper is the following idea:

\textit{Combining techniques from model-based RL with human preference learning offers several advantages:
    (1) safer and more sample efficient reward learning and policy optimization,
    (2) diverse trajectories useful for preference learning, which are a free byproduct of model-based RL, and
    (3) reward pre-training with automatically generated preferences over demonstrations without any environment interaction.
}

\begin{figure*}
    \centering
    \includegraphics[width=0.92\textwidth]{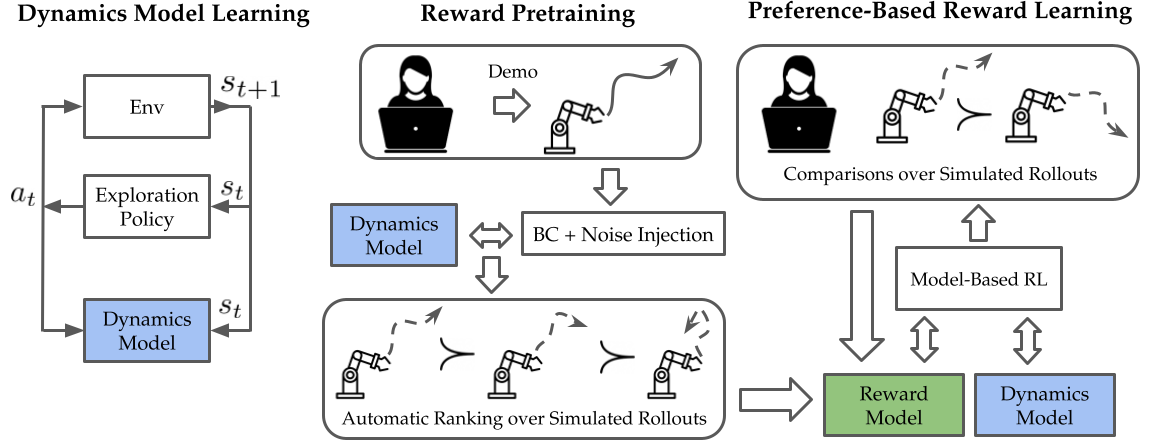}
    \caption{\textbf{Model-Based Preference-Based RL (MoP-RL):}
    We leverage intrinsic exploration (left) to learn a dynamics model that enables efficient reward pre-training (center) by running behavioral cloning (BC) on a small number of demonstrations and using noise injection to simulate a variety of worse trajectories using rollouts in the learned dynamics model. We fine-tune the learned reward function via active preference queries to a human (right), where the trajectories for queries are simulated using the learned dynamics model and produced as a byproduct of running model-based RL with the learned reward function. \alg uses the learned dynamics model to enable safe and sample efficient learning of human preferences.}
    \label{fig:pipeline}
\end{figure*}

\newsec{Safe and sample efficient learning} 
Combining model-based RL with PbRL allows robots to use a learned dynamics model to hypothesize different behaviors and even show these planned behaviors to a human without executing them in the environment, while still learning customizable behaviors. This enables safe and efficient reward function learning and policy optimization by limiting the number of environmental interactions. Though learning a dynamics model may require a large amount of data, we may amortize this cost by learning multiple reward functions for different users via the same learned dynamics model.

\newsec{Diverse trajectories for preference learning}
A learned dynamics model can also be used to simulate diverse trajectories so that preference queries can be created without any environment interaction. While such query generation has safety and efficiency benefits, since it requires no environment interaction, one must still decide which trajectories to show the human. We leverage the fact that, by design, many model-based RL approaches already generate a diverse set of trajectories. Given the learned dynamics model, we use model predictive control (MPC)~\cite{rossiter2017model} based on the cross entropy method (CEM)~\cite{botev2013cross} to optimize trajectories. This common MPC approach uses a cost function (learned from preferences in our case) to iteratively refine an action sequence~\cite{williams2015model,finn2017deep,chua2018deep,ebert2018visual,hafner2019learning,nagabandi2020deep}. Our insight is that this allows us to generate diverse preference queries without any extra computation---trajectories generated by successive CEM iterations provide a diverse set of trajectories that can be reused for preference learning. Because successive CEM iterations converge to the model's belief of the human's preferences, obtaining preference queries over successive CEM iterations also allows the supervisor to quickly correct the learned reward function if it leads to undesired behavior.

\newsec{Reward pre-training from suboptimal demonstrations}
Prior work has proposed Disturbance-based Reward Extrapolation (D-REX)~\cite{brown2020better} as a way to automatically produce ranked trajectories by injecting noise into a policy learned via behavioral cloning. While this approach can lead to well-shaped reward functions and enable better-than-demonstrator performance, it requires rolling out noisy trajectories in the environment, which can be unsafe and slow. However, with a learned dynamics model, we can perform such noisy rollouts within the learned model. We study this approach and show that we can leverage a small number of suboptimal demonstrations to achieve improvements in preference learning using model-based reward pre-training without requiring any environmental interactions.

In summary, our work makes the following contributions:
\begin{enumerate}
    \item The first analysis of preference-based RL with a learned dynamics model (see Figure~\ref{fig:pipeline}).
    \item A novel preference-generation approach that uses the samples produced as a byproduct of model-based RL to generate diverse and informative preference-queries without requiring physical rollouts in the environment.
    \item An approach for reward pre-training together with \alg, which enables combining suboptimal demonstrations and pairwise preferences in a model-based setting.
    \item Experiments suggesting that our approach can learn to perform complex tasks from preferences with fewer environmental interactions than prior approaches and can scale to high-dimensional visual control tasks.
\end{enumerate}


\section{RELATED WORK}

\newsec{Model-based learning}
There has been much interest in learning dynamics models~\cite{nguyen2011model,grunewalder2012modelling,thuruthel2017learning,kolter2019learning} and applying these learned models to RL~\cite{chua2018deep,okada2020variational,zhang2020asynchronous,veerapaneni2020entity,yu2020mopo,wang2021rough,bechtle2020curious,wu2021example,xu2022look, lowrey2018plan}. Past work has indicated that model-based RL can achieve higher sample efficiency than model-free methods~\cite{kaiser2019model, zhang2019solar}.
Outside of RL, learned dynamics models have also been used successfully in imitation learning methods that learn from expert demonstrations. For instance, forward and inverse dynamics models have been shown to be useful when learning from demonstrations consisting of state-action pairs~\cite{finn2016guided,baram2016model, baram2017end,wu2020model}, as well as when learning from observations without access to the demonstrator's actions~\cite{torabi2018behavioral,edwards2019imitating,kidambi2021mobile}. 
ReQueST~\cite{reddy2020learning} actively collects human reward labels over model-based trajectory rollouts and then performs model-based RL using a learned reward function. However, our work differs in the following important ways: (1) ReQueST only performs model-based RL on the learned rewards after completion of learning; in contrast, \alg leverages the reward learned \textit{so far} to synthesize human feedback queries using informed trajectories that improve throughout learning. (2) ReQueST requires a human to provide a reward label to every state in each queried trajectory, resulting in a significantly higher human burden than \alg, which requires only pairwise preferences over entire trajectories.

\newsec{Preference-based RL} It is often easier for a person to qualitatively rank two or more behaviors than to demonstrate a good behavior.
Learning from pairwise preferences over trajectories is a common approach to learning reward functions and corresponding RL policies. However, despite the widespread interest in preference-based RL, most prior work either takes a model-free approach~\cite{wirth2016model,christiano2017deep,ibarz2018reward,brown2020safe,lee2021pebble} or assumes that the system dynamics are perfectly known~\cite{zucker2010optimization,wilson2012bayesian,gritsenko2014learning,wirth2014learning,jain2015learning,sadigh2017active}. By contrast, we seek to study preference-based RL when using a learned dynamics model.
Prior works on model-free preference-based RL demonstrate the effectiveness of pairwise preference learning for a range of RL tasks~\cite{christiano2017deep,brown2019extrapolating,lee2021pebble,tien2023study}. In particular, PEBBLE~\cite{lee2021pebble} achieves state-of-the-art performance of PbRL in simulated robot locomotion and manipulation tasks. However, model-free RL tends to require more environment interaction than model-based RL methods~\cite{kaiser2019model, zhang2019solar}, and we are not aware of prior work that studies PbRL with dynamics model learning. Shin et al.~\cite{shinbenchmarks} study PbRL for offline RL which avoids online interactions, but requires a large dataset of prior trajectories.
Novoseller et al.~\cite{novoseller2020dueling} learn a dynamics model and a reward predictor from human preferences in tabular environments, whereas our experiments focus on continuous, higher-dimensional environments. 
Our experiments suggest that \alg achieves better policy performance than model-free PbRL with significantly fewer environment interactions. 

Our work uses noise injection into the learned model for reward pretraining. Recent work by Chen et al.~\cite{chen2021learning} provides evidence of inductive bias when pre-training using noise injection and a preference-learning objective, as done in our paper. However, Chen et al.'s approach requires solving a sequence of optimization problems and running online IRL, and thus requires a large amount of computation time and many on-policy rollouts~\cite{chen2021learning}. 
We leave it to future work to investigate alternatives to D-REX~\cite{brown2020better} for reward pre-training without environmental interaction.



\begin{figure*}
    \centering
    \includegraphics[width=0.93\textwidth]{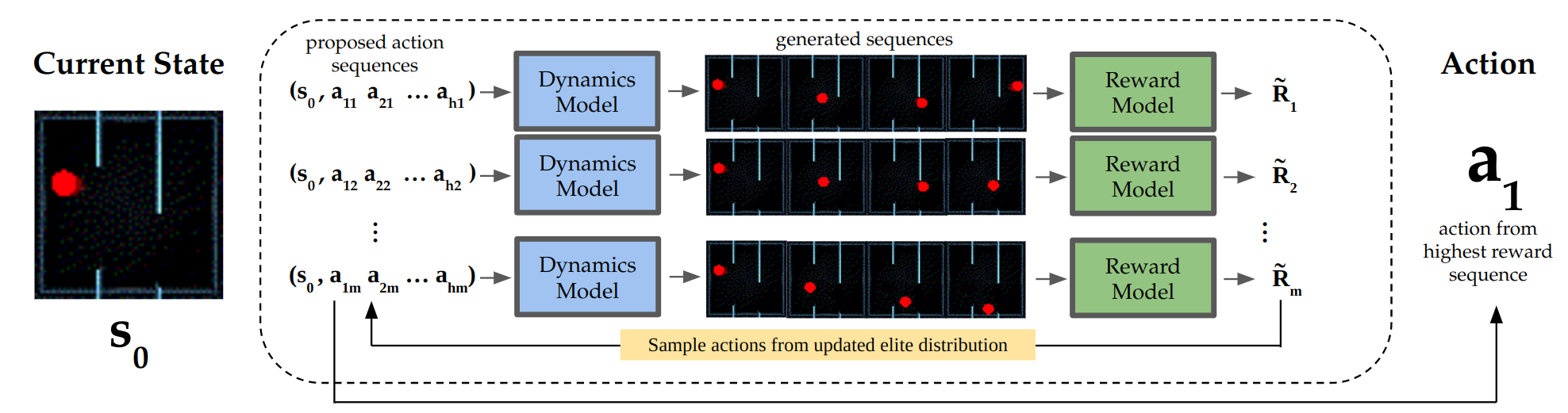}
    \caption{\textbf{Cross Entropy Method:} For an initial state $s_0$, we sample a set of action sequences and predict the corresponding state trajectory for each one using the dynamics model $\hat{f}_{\phi}$. We estimate rewards for these trajectories as a linear combination of the reward model $\hat{r}_{\theta}$'s prediction and an RND bonus. The $m_e$ action sequences with the highest predicted rewards form the elites; these are used to update the CEM distribution and sample the next population of action sequences. In the final CEM iteration, the mean of the action distribution of elites is selected for execution in the environment. }
    \label{fig:CEM_pipeline}
\end{figure*}

\section{PROBLEM DEFINITION}


\newsec{Problem formulation and notation}
We formulate the problem as a Markov Decision Process (MDP), $\mathcal{M} = (\mathcal{S}, \mathcal{A}, P, r)$, where $\mathcal{S}$ is the state space, $\mathcal{A}$ is the action space, $P: \mathcal{S} \times \mathcal{A} \times \mathcal{S} \to [0, 1]$ specifies the state transition probabilities, and $r$ is the human's reward function, which, importantly, we assume is unobserved by the learning agent.  A robot \textit{trajectory} takes the form $\tau = (s_0, a_0, s_1, a_1, \ldots, s_T)$ for some episode length $T$.
The robot does not observe numerical rewards. Rather, we assume access to a human user who can answer pairwise preference queries of the form, ``Do you prefer trajectory A or B?'' as well as possibly provide a small set $\mathcal{D}_{\rm demo}$ of suboptimal demonstrations, where $\mathcal{D}_{\rm demo}$ consists of trajectories $\tau^E = (s_0^E, a_0^E, \ldots, s_T^E) \in \mathcal{D}_{\rm demo}$.
The pairwise preference $\tau_A \succ \tau_B$ indicates that the user states a preference for trajectory $\tau_A$ over $\tau_B$.
We assume that the human's preferences are based on a true but unobserved reward function, $r: \mathcal{S} \times \mathcal{A} \to \mathbb{R}$, which quantifies the utility of each state-action pair to the human. The total underlying reward of a trajectory $\tau = (s_0, a_0, s_1, a_1, \ldots, s_T)$ is denoted via $J(\tau) := \sum_{t=0}^{T - 1} r(s_t, a_t)$.
A \textit{policy} $\pi: \mathcal{S} \times \mathcal{A} \to [0, 1]$ maps states to distributions over actions. We aim to learn a policy  $\pi$ that matches user preferences by maximizing expected performance under the human's true reward function $r$. Our only access to $r$ is through the demonstrations and pairwise preferences. 



\newsec{Assumptions} We assume access to a human who can give preferences over trajectories based on their internal utility function as described above. As part of this assumption, we assume that we can visualize trajectories for the human to rank. Note that we only assume access to a method to show the human a visualization of states using the learned dynamics model---we do not assume to know or have a model of the true system dynamics.

\newsec{Performance metrics} We aim to identify the optimal policy $\pi^*$ that maximizes the human's total underlying reward in expectation:
$
    \pi^* = \text{argmax}_{\pi} \mathbb{E}_{\tau \sim \pi} [J(\tau)] = \text{argmax}_{\pi} \mathbb{E}_{\tau \sim \pi}\left[\sum_{(s,a) \in \tau} r(s,a) \right].
$
Furthermore, we aim to discover $\pi^*$ while minimizing the total agent-environment interaction and the total number of human interactions.


\section{METHODS}

\subsection{Learning Rewards from Preferences}
To obtain a well-performing policy $\pi$, we learn a predictor $\hat{r}_{\theta}: \mathcal{S} \times \mathcal{A} \to \mathbb{R}$ of the user's reward $r$, parameterized by $\theta$. We model the probability $\hat{P}$ that the user prefers trajectory $\tau_1$ to $\tau_2$ via the Bradley-Terry model~\cite{brown2019extrapolating, christiano2017deep}:
$
    \hat{P}(\tau_1 \succ \tau_2) = \frac{\exp(\hat{J}_{\theta}(\tau_1))}{\exp(\hat{J}_{\theta}(\tau_1)) + \exp(\hat{J}_{\theta}(\tau_2))},
$ 
where $\hat{J}_{\theta}(\tau) := \sum_{s, a \in \tau}\hat{r}_{\theta}(s, a)$. Given a collection of user preferences $\mathcal{D}_{\text{pref}}$, we learn $\theta$ by minimizing the following cross-entropy loss function~\cite{christiano2017deep, brown2019extrapolating, brown2020better}: 
\begin{equation}\label{eqn:pref}
\mathcal{L}_{\rm pref}(\theta) = -\sum_{\tau_i \succ \tau_j \in \mathcal{D}_{\text{pref}}}  \log \left[\frac{\exp{\left(\hat{J}_{\theta}(\tau_i)\right)}}{\exp{\left(\hat{J}_{\theta}(\tau_i)\right)} + \exp{\left(\hat{J}_{\theta}(\tau_j)\right)}} \right].
\end{equation}

With image observations, rather than directly predicting rewards via $\hat{r}_{\theta}(s, a)$, we use an LSTM to generate reward predictions that integrate over a sequence of observations. The LSTM takes in a state sequence, and its final output is fed into the feedforward network $\hat{r}_{\theta}$ to predict the reward.


\subsection{Learning a Dynamics Model}

Given a dataset of observed state-action transitions, $\mathcal{D}_{\text{dyn}} = \{(s_i, a_i, s_{i+1})\}_{i = 1}^M$, we learn a transition dynamics model $\hat{f}_{\phi}: \mathcal{S} \times \mathcal{A} \to \mathcal{S}$, parametrized by $\phi$. We learn $\phi$ by minimizing the following loss:
$
    \label{eqn:dyn_loss}
    \mathcal{L}_{\text{dyn}}(\phi) := \sum_{(s, a, s_{\text{next}}) \in \mathcal{D}_{\text{dyn}}} ( s_{\text{next}} - \hat{f}_{\phi}(s, a) )^2 + \lambda||\phi||^2,
$
where our forward dynamics network predicts the change in state, $\hat{f}^{\rm diff}_{\phi}(s, a) := \hat{f}_{\phi}(s, a) - s$, and $\lambda$ is a weight decay hyperparameter. We pre-process the data by normalizing it to have zero mean and unit variance along each dimension. 
When the observations are images, we adapt Stochastic Video Generation with a Learned Prior (SVG-LP) \cite{denton2018stochastic} to train a visual dynamics model.

This work leverages two methods for collecting the dynamics training dataset $\mathcal{D}_{\text{dyn}}$. The first is random agent-environment interaction, in which the agent samples its action space uniformly randomly over a series of trajectories beginning in different start states. However, in many domains, random agent-environment interaction may yield insufficient exploration.
To collect diverse data for learning a dynamics model, we leverage random network distillation (RND)~\cite{burda2019exploration}, a powerful approach for exploration in deep RL that provides reward bonuses based on the error of a neural network predicting the output of a randomly-initialized network. 
We gather a dataset of state-action transitions observed while training a Soft Actor Critic~\cite{haarnoja2018soft} policy using the RND bonus as the sole reward signal. In addition to using RND for dynamics pre-training, we integrate RND into our online learning pipeline as discussed
in Section~\ref{ssec:mbrl}.

\subsection{Pre-training with Suboptimal Demonstrations}
Brown et al.~\cite{brown2020better} propose the D-REX algorithm, in which demonstration trajectories are automatically ranked based on the degree of noise used to generate them. 
We adapt D-REX to the model-based setting. Given a set of human demonstrations $\mathcal{D}_{\text{demo}}$, we train a behavior cloning policy $\pi_{BC}: \mathcal{S} \to \mathcal{A}$ on each state-action pair in $\mathcal{D}_{\text{demo}}$. Importantly, we roll out this policy inside the \textit{learned dynamics model}. Simulated rollouts are generated with $\epsilon$-greedy noise, such that the agent takes a uniformly random action with probability $\epsilon$ and follows $\pi_{BC}$ otherwise. For trajectories $\tau_i$ and $\tau_j$ starting from the same initial state and generated with noise levels $\epsilon_i$ and $\epsilon_j$, $\epsilon_i > \epsilon_j$, we automatically rank $\tau_j \succ \tau_i$. We also prefer any trajectory $\tau \in \mathcal{D}_{\text{demo}}$ over any trajectory generated by rolling out $\pi_{BC}$. 

\subsection{Model-Based RL}\label{ssec:mbrl}
We adopt a model-based RL approach inspired by Chua et al.~\cite{chua2018deep}, leveraging the cross-entropy method (CEM)~\cite{rubinstein1999cross, rubinstein2004cross} to plan with respect to learned reward and dynamics models (see Fig.~\ref{fig:CEM_pipeline}). CEM maintains a distribution $\mathcal{N}(\bm{\mu}_i, \Sigma_i)$ in each CEM iteration $i$, from which the $i$\textsuperscript{th} population of $m$ action sequences is sampled: $\{a_{1j}^{(i)}, \ldots, a_{hj}^{(i)} \mid j=1, \ldots, m\}$, with planning horizon $h$. 
We then use the learned dynamics and rewards, $\hat{f}_{\phi}$ and $\hat{r}_{\theta}$, to estimate the expected total reward $\tilde{R}_j^{(i)}$ of each action sequence, conditioned on a fixed start state $s_0 \in \mathcal{S}$: $\hat{s}_{1j}^{(i)} = \hat{f}_{\phi}(s_0, a_{1j}^{(i)})$, $\hat{s}_{2j}^{(i)} = \hat{f}_{\phi}(\hat{s}_{1j}^{(i)}, a_{2j}^{(i)})$, 
etc., and $\tilde{R}_j^{(i)} = \sum_{t=0}^{h-1} \hat{r}_{\theta} (\hat{s}_{tj}^{(i)}, a_{(t+1)j}^{(i)})$, 
where $\hat{s}_{0j}^{(i)} := s_0$.
The $m$ population members are then ranked according to their estimated rewards $\tilde{R}_j^{(i)}$, and the top $m_e < m$ population members are termed the \textit{elites}. We take the mean and variance of the elites to refine the CEM distribution, obtaining $\mathcal{N}(\bm{\mu}_{i + 1}, \Sigma_{i + 1})$.
To choose actions in the environment, we employ model predictive control (MPC), as in~\cite{chua2018deep, lowrey2018plan}: we execute the first $k$ actions of the action sequence given by $\bm{\mu}_{\rm final}$, the mean of the action distribution calculated in the final CEM iteration. Our experiments utilize $k = 1$.


We additionally leverage RND~\cite{burda2019exploration} to improve exploration. RND provides a prediction error signal $\hat{g}_{\rho}(s)$ at each state $s \in \mathcal{S}$, parameterized by $\bm{\rho}$. Intuitively, areas with higher error $\hat{g}_{\rho}(s)$ have been explored less, and so $\hat{g}_{\rho}(s)$ functions as an exploration bonus. During CEM, we augment the expected reward: $\tilde{R}_j^{(i)} = \sum_{t=0}^{h-1} [\hat{r}_{\theta}(\hat{s}_{tj}^{(i)}, a_{(t+1)j}^{(i)}) + \lambda_{RND} \hat{g}_{\rho}(\hat{s}_{tj}^{(i)})]$, where $\lambda_{RND}$ is a hyperparameter.

\subsection{Active Model-Based Preference Query Generation}
To efficiently generate pairwise preference queries without additional environment interaction, we form queries using trajectories from the CEM process, during which the learned dynamics model $\hat{f}_{\phi}$ is used to roll out candidate action sequences. In selecting pairs of CEM trajectories to elicit preferences, we consider two main questions: (1) How can we generate a diverse pool $\mathcal{D}_{\text{traj}}$ of trajectories, e.g., so that the human is not forced to give preferences between overly-similar or otherwise limited choices? (2) Given a pool $\mathcal{D}_{\text{traj}}$ of candidate trajectories, how can we select which trajectory pairs are best to show to the human?

To address (1), we include trajectories from all CEM iterations in $\mathcal{D}_{\text{traj}}$. This ensures that $\mathcal{D}_{\text{traj}}$ contains diverse data, including from non-elite trajectories. Notably, constructing preference queries using only the final CEM iteration could force the human to compare overly-similar trajectories, leading to unreliable preference labels. At a randomly-selected time $t_{keep} \in \{1, \ldots, T\}$ in each episode, we save $m_{\rm keep} < m$ trajectories from each CEM iteration in $\mathcal{D}_{\text{traj}}$; these are randomly selected from the elites in the final CEM iteration and from among non-elites in previous iterations. 
We randomly choose a new $t_{\rm keep}$ in each episode so that across episodes, $\mathcal{D}_{\text{pref}}$ contains pairs of trajectories beginning in different states. For (2), we maximize the information gain~\cite{erdem2020asking, biyik2020active} to identify trajectory pairs in $\mathcal{D}_{\text{traj}}$ that are expected to be most informative about the human's underlying reward $r$. 
We identify the $N$ trajectory pairs $\{\tau_1^q, \tau_2^q \mid \tau_1^q, \tau_2^q \in \mathcal{D}_{\text{traj}}, q = 1, \ldots, N\}$ with the highest mutual information $I(r; \mathbb{I}_{[\tau_1^q \succ \tau_2^q]} \mid \mathcal{D}_{\text{pref}})$ between their preference outcome and the unknown reward $r$~\cite{erdem2020asking, biyik2020active}:
\begin{equation}\label{eqn:info_gain}
\begin{split}
    I(r; \mathbb{I}_{[\tau_1 \succ \tau_2]} \mid \mathcal{D}_{\text{pref}}) &= H(\mathbb{I}_{[\tau_1 \succ \tau_2]} \mid \mathcal{D}_{\text{pref}}) - \\ & \mathbb{E}_{r \sim p(r \mid \mathcal{D}_{\text{pref}})}[H(\mathbb{I}_{[\tau_1 \succ \tau_2]} \mid r, \mathcal{D}_{\text{pref}})],
\end{split}
\end{equation}
where $\mathbb{I}_{[\tau_1 \succ \tau_2]} \in \{0, 1\}$ indicates the outcome of a preference query between $\tau_1$ and $\tau_2$, $\mathcal{D}_{\text{pref}}$ is the pairwise preference dataset (so far), $H$ is information entropy, and we use an ensemble of reward networks to estimate the posterior $p(r \mid \mathcal{D}_{\text{pref}})$.
Algorithm 1 details the entire \alg algorithm.

\setlength{\textfloatsep}{2mm}
\begin{algorithm}[t]
    \label{alg:algorithm}
    \SetKwInOut{Input}{Input}
    \SetKwInOut{Output}{Output}
    \SetKwInOut{Parameters}{Parameters}
    \SetKwInOut{Subroutines}{Subroutines}
    Initialize parameters $\theta, \phi$ of reward and dynamics networks $\hat{r}_\theta$ and $\hat{f}_\phi$, respectively \\
    Collect dataset $\mathcal{D}_{\rm dyn}$ of transitions in environment using exploration strategy (random or RND) \\
    Optimize $\mathcal{L}_{\rm dyn}$ in \eqref{eqn:dyn_loss} with respect to $\phi$\\
    Initialize $\mathcal{D}_{\rm pref} \gets \emptyset$ \\
    Collect dataset $\mathcal{D}_{\rm demo}$ of human demonstrations \\
    Initialize $\hat{r}_\theta$ via model-based D-REX using $\mathcal{D}_{\rm demo}$ and using $\hat{f}_\phi$ to roll out trajectories \\
    \For{each episode $k$}{
        Initialize $\mathcal{D}_{\rm traj} \gets \emptyset$ \\
        $t_{\rm keep} \sim \mathcal{U}\{1, \ldots, T\}$ \\
        \For{each timestep $t \in \{1, \ldots, T\}$}{
            $s_t \gets $ current state \\ 
            Select optimal action sequence $(a_t, \ldots, a_{t+h-1})$ using CEM with $\hat{f}_\phi$, $\hat{r}_\theta$, and $\hat{g}_{\rho}$ \\
            Execute $a_t$ in the environment
            
            \If{$t = t_{\rm keep}$}{
                $\mathcal{D}_{\rm traj} \gets  m_{\rm keep}$ trajectories from each CEM iteration
            }
        }
        $\{\tau_1^{q}, \tau_2^q \mid q = 1, \ldots, N\} \gets $ top $N$ trajectory pairs in $\mathcal{D}_{\rm traj}$ according to info-gain \eqref{eqn:info_gain} \\
        \For{$q = 1, \ldots, N$ }{
            Query human for preference $y$ \\
            $\mathcal{D}_{\rm pref} \gets \mathcal{D}_{\rm pref} \cup \{\left( \tau_1^q, \tau_2^q, y\right) \}$ \\
        }
        \If{$k$ \% reward update frequency = 0}{
        Optimize $\mathcal{L}_{\rm pref}$ in \eqref{eqn:pref} with respect to $\theta$ \\
        }
        \If{$k$ \% RND update frequency = 0}{
        Update RND network $g_{\rho}$ on visited states
        }
     }
    \caption{\alglong (\alg)}
\end{algorithm}
\setlength{\floatsep}{2mm}


\section{EXPERIMENT RESULTS}
\label{sec:result}

\subsection{Experiment Domains}

Our experiments consider four simulation domains detailed below. 
For all domains, preferences are provided by a synthetic labeler that does not make mistakes. To simulate the fact that humans have difficulty answering some queries~\cite{erdem2020asking}, the labeler skips preference queries where the reward difference falls below a threshold. Additional experiment details are provided in the supplementary website.

\newsec{Maze-LowDim} The agent must navigate a 2-dimensional maze to reach a goal location. The agent's state is given by its $(x, y)$ coordinates in the maze. The agent's action is a vector $(f_x, f_y) \in \mathbb{R}^2$, $-f_{\max} \le f_x, f_y \le f_{\max}$, representing a force applied in each coordinate direction. 
The synthetic labeler provides preferences using a hand-specified underlying cost function, which gives higher preference first for reaching the far right wall and then the top-right corner.

\newsec{Maze-Image} This environment uses the same maze, agent dynamics, and synthetic reward as Maze-LowDim; however, the agent observes an image of the environment (as in Fig.~\ref{fig:CEM_pipeline}) at each step. 

\begin{figure}
    \centering
    \begin{subfigure}[b]{0.49\linewidth}
         \includegraphics[width=0.9\linewidth]{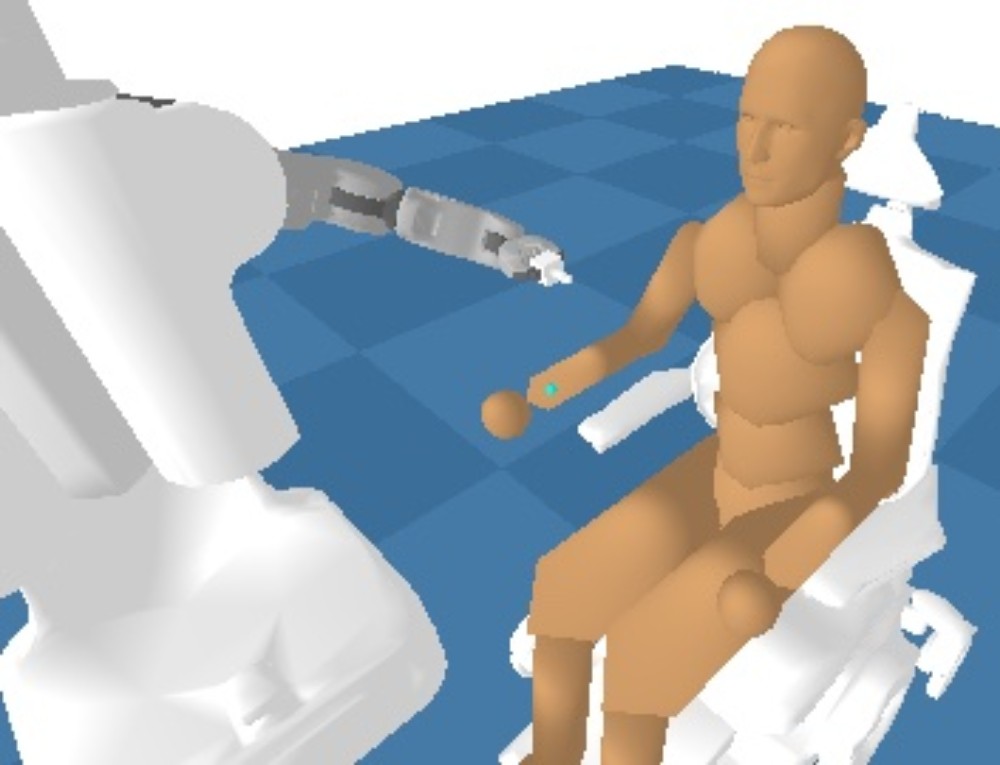}
     \end{subfigure}
    \hfill
     \begin{subfigure}[b]{0.49\linewidth}
         \includegraphics[width=0.9\linewidth]{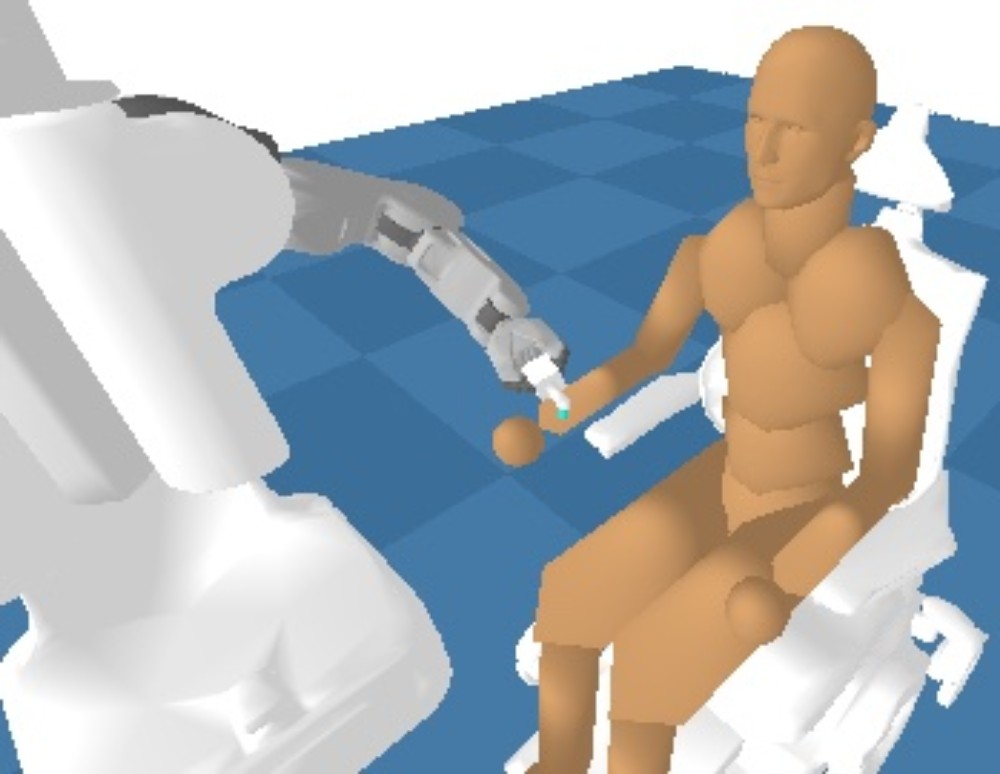}
     \end{subfigure}
    \caption{Assistive Gym start (left) and goal (right) state.}
\label{fig:AG}
\end{figure}

\newsec{Assistive-Gym}
We utilize a simplified version of the itch scratching task in the Assistive Gym~\cite{erickson2020assistive} simulation environment (Fig.~\ref{fig:AG}), in which a robot manipulator must position itself on an itch on a person's arm and apply a target amount of force. We use a reduced state space containing 1) the vector difference between the robot end effector and itch location and 2) the amount of force applied by the end effector. The synthetic labeler prefers trajectories that bring the end effector close to the itch and apply an amount of force below a threshold to the person's arm.

\newsec{Hopper}
In the OpenAI Gym~\cite{gym} Hopper environment, we aim to train the agent to perform a backflip. The state space $\mathcal{S} \subset \mathbb{R}^{11}$ consists of the angular positions and velocities of the Hopper's 3 joints and the top of the robot, as well as the height of the hopper and the velocities of the x-coordinate and z-coordinate of the top. The action space $\mathcal{A} \subset \mathbb{R}^3$ consists of a torque applied to each of the 3 joints.
Pairwise preferences over trajectories are given by a synthetic labeler with a hand-engineered reward function designed by Christiano et al.~\cite{christiano2017deep}.

\subsection{Performance Metrics and Algorithms Compared}

Our experiments compare the performance of~\alg to PEBBLE~\cite{lee2021pebble}, a state-of-the-art model-free PbRL algorithm.  For each simulation domain, all algorithms learn from the same number of pairwise preferences.
Performance is evaluated via the following metrics:
\begin{enumerate}
    \item  \textbf{Success rate} (Maze-LowDim, Maze-Image): percentage of times that the agent satisfies a goal condition.
    \item \textbf{Average reward} (Assistive-Gym): ground truth reward over an entire episode, averaged over several rollouts.
\end{enumerate}

\begin{figure*}[t!]
    \centering
    \includegraphics[width=\textwidth]{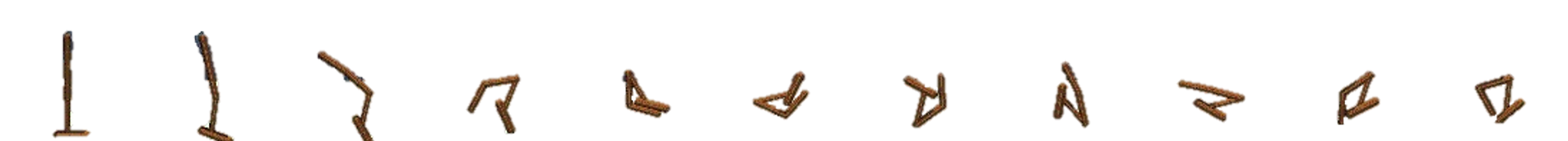}
    \caption{\alg trains a Hopper to perform a backflip via preference queries over learned dynamics.}
    \label{fig:hopper}
\end{figure*}








In each environment,~\alg first trains a dynamics model. In Maze-LowDim, we collect dynamics training data via random environment interaction, while in the other domains---in which exploration is more challenging---we collect dynamics data via RND. Then,~\alg performs interactive reward learning, possibly with reward pre-training from demonstrations. The reward model $\hat{r}_{\theta}$ is a function of state and action in Assistive-Gym and Hopper, and of state only in both maze environments. PEBBLE performs an initial unsupervised exploration phase prior to reward learning.
\alg used 1000 trajectories to learn the dynamics model in Maze-LowDim, 2000 in Maze-Image, 2500 for Assistive-Gym, and 5000 for Hopper. PEBBLE used 150 trajectories in the unsupervised step in Maze-LowDim, 2000 or 5000 in Maze-Image (see Table~\ref{table:maze_image_performance}), and 150 in Assistive-Gym. 



\begin{table}[t]
\centering
\begin{tabular}{l r r }
 \hline
 Algorithm &  Training Rollouts & Success Rate\\
 \hline
 PEBBLE   & 50 & 6.3\% \\
 PEBBLE   & 100 &  81.3\% \\
 PEBBLE   & 200 &  100\% \\
 \hline
 \alg (w/o demos)   & 18 & 0.0\% \\
 \alg (w/o demos)   & 24 & 56.3\% \\
 \alg (w/o demos)& 30 & 100\%\\
 \hline
 \alg (w/ 2 demos) & 18 & 100\% \\
 \alg (w/ 2 demos)   & 24 & 100\% \\
 \alg (w/ 2 demos)  & 30 & 100\% \\
 \hline
\end{tabular}
\caption{\textbf{Performance on Maze-LowDim Environment.} We report the numbers of unsupervised and training rollouts and the success rate (over 16 rollouts) for each method when trained with 60 preference queries.}
\label{table:maze_lowdimperformance}
\end{table}

\begin{table}[t]
\centering
\begin{tabular}{ p{2.7cm} p{1.3cm} p{1.2cm} p{1.1cm} p{1cm}  }
 \hline
 Algorithm & Unsupervised Rollouts & Training Rollouts & Success Rate\\
 \hline
 PEBBLE   & 2000 & 3300  & 0.0\% \\
 PEBBLE   & 5000 & 3300  & 0.0\% \\
 \hline
 \alg (w/o demos)  & 2000 & 800  & 25.0\% \\
 \hline
 \alg (w/ 10 demos)  & 2000 & 800  & 68.8\% \\
 \hline
\end{tabular}
\caption{\textbf{Performance on Maze-Image Environment.} We report the numbers of unsupervised and training rollouts and the success rate (over 16 rollouts) for each method compared. Both methods were trained with 1000 preference queries.}
\label{table:maze_image_performance}
\end{table}

\begin{table}[t]
\centering
\begin{tabular}{l r r}
 \hline
 Algorithm & Training Rollouts & Average Reward \\
 \hline
 PEBBLE   &  100 & -46.1 $\pm$ 1.36\\
 \hline
 \alg   &  8 & -26.9 $\pm$ 6.82\\
 \hline
\end{tabular}
\caption{\textbf{Performance on Assistive-Gym Itch Scratching Environment.} We report the numbers of unsupervised and training rollouts and the average reward (mean $\pm$ standard error over 4 rollouts) for each method compared. Both methods were trained with 80 preference queries.}
\label{table:itchperformance}
\end{table}








\subsection{Results}\label{ssec:results}

We hypothesize that~\alg will require fewer environmental interactions in comparison to model-free RL, while achieving similar or better performance. We compare~\alg and PEBBLE~\cite{lee2021pebble} in the Maze-LowDim, Maze-Image, and Assistive-Gym Itch Scratching tasks.

\newsec{Maze-LowDim} Table~\ref{table:maze_lowdimperformance} shows results for the Maze-LowDim environment. We see that PEBBLE requires a large number of rollouts to learn the reward function accurately enough to successfully navigate the maze. By contrast, \alg is much more sample efficient. 
We also observe that pre-training the reward network using two demonstrations significantly speeds up online preference learning, requiring fewer trajectories to successfully complete the task.

\newsec{Maze-Image} Table~\ref{table:maze_image_performance} shows results for the Maze-Image environment. We see that \alg outperforms PEBBLE, even when PEBBLE is given significantly more unsupervised access to the environment. Prior work~\cite{lee2021pebble} only evaluates PEBBLE on low-dimensional reward learning tasks. We adapted the author's implementation of PEBBLE to allow an image-based reward function and use the same reward function architecture for both PEBBLE and \alg. However, our results demonstrate that PEBBLE struggles in high-dimensional visual domains. Furthermore, we see that pre-training with 10 suboptimal demonstrations significantly improves reward learning without any additional environment interactions. Pretraining leads to improved task success by first learning a rough estimate of the reward function and then fine-tuning via model-based preference queries. 

\newsec{Assistive Gym} Table~\ref{table:itchperformance} shows the results for the itching task from Assistive Gym. \alg also achieves a higher task reward than PEBBLE in this environment.

The above experiments demonstrate that~\alg requires significantly fewer environment interaction steps than PEBBLE to learn a reward function. \alg also enables dynamics model pre-training to be performed separately from preference learning. This is important since, unlike a learned reward function, learned dynamics can be re-used to safely and efficiently learn the preferences of multiple users.

\newsec{Hopper Backflip} Inspired by previous model-free PbRL results~\cite{christiano2017deep}, we demonstrate that~\alg can train the OpenAI Gym Hopper to perform a backflip via preference queries over a learned dynamics model.
An example learned backflip, displayed in Figure~\ref{fig:hopper}, suggests that~\alg can learn to perform novel behaviors for which designing a hand-crafted reward function is difficult.



\section{CONCLUSION}
\label{sec:conclusion}

This work introduces~\alg, a model-based approach to preference-based RL. We demonstrate that dynamics modeling is uniquely suited for preference-based learning, since a learned dynamics model can be used to generate pairwise preference queries without environment interaction and to greatly reduce the amount of environmental interaction needed to optimize a policy from the learned reward function.
Furthermore, \alg combines multiple human interaction modalities by integrating preference learning with model-based reward pre-training from demonstrations for improved performance.
Limitations include that we synthetically generated all pairwise preferences and that we do not consider noisy preference data.
Future work includes evaluating~\alg in a user study, investigating safety-critical domains, and exploring applications of~\alg, in environment where interactions are expensive and where different human users have varying robot interaction preferences.







\renewcommand*{\bibfont}{\footnotesize}
\printbibliography

\end{document}